\documentclass{article}

\usepackage{xcolor}

\usepackage[preprint]{corl_2026}

\usepackage[T1]{fontenc}
\usepackage[utf8]{inputenc}
\usepackage{amsmath,amssymb,amsfonts}
\usepackage{bbm}
\usepackage{graphicx}
\usepackage{booktabs}
\usepackage{url}
\usepackage{algorithmic}
\usepackage[ruled,vlined,linesnumbered]{algorithm2e}
\usepackage{multirow}
\usepackage{array}
\usepackage{wrapfig}

\usepackage{tikz}
\usetikzlibrary{matrix}

\usepackage{hyperref}

\usepackage{tikz}
\usetikzlibrary{matrix}

\newcommand{\ACTrgb}{ACT\,(RGB)}
\newcommand{\ACTtob}{ACT\,(TOB)}
\newcommand{\DPrgb}{DP\,(RGB)}
\newcommand{\DPrgbd}{DP\,(RGB\nobreakdash-D)}
\newcommand{\DPtob}{DP\,(TOB)}
\newcommand{\Ours}{ObstaDiff}

\title{ObstaDiff: Generalizable Diffusion Policy Learning via Obstacle-aware Representations}

\author{
  Jiawen Wang$^{1}$ \quad
  Kevin Yao$^{2}$ \quad
  Khalid Jawed$^{1,*}$\\
  $^{1}$Department of Mechanical and Aerospace Engineering,
  University of California, Los Angeles\\
  $^{2}$Department of Computer Science,
  University of California, Los Angeles\\
  $^{*}$Corresponding author:
  \texttt{khalidjm@seas.ucla.edu}
}

\begin{document}
\maketitle

\begin{figure}[ht]
    \centering
    \vspace{-0.2in}
    \includegraphics[width=\linewidth]{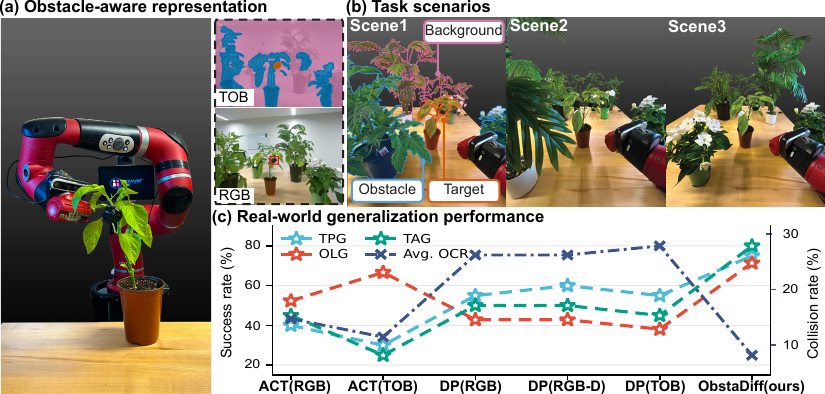} 
\caption{\textbf{\Ours{}} learns a structured target--obstacle--background (TOB) representation
for imitation learning in cluttered scenes. This representation enables
obstacle-aware approach motions, leading to \textbf{improved generalization} across
target-pose, obstacle-layout, and target-appearance settings while \textbf{reducing the
average obstacle collision rate}, as shown in (c).
}
    \label{fig:teaser}
    \vspace{-0.1in}
\end{figure}

\begin{abstract}
Imitation learning has achieved impressive results in robotic manipulation, yet
most existing approaches assume clean backgrounds and lack explicit mechanisms
for obstacle-aware motion generation. Extending such policies to cluttered,
real-world scenes with unstructured obstacles remains a key generalization
challenge. We present \textbf{\Ours{}}, a decomposed diffusion-policy
framework with a lightweight \textit{obstacle-aware visual encoder}. \Ours{}
extracts a structured \emph{target--obstacle--background} representation,
enabling the downstream alignment policy to generate end-effector trajectories
toward a target-centered bottleneck pose while reasoning about surrounding
obstacles. We evaluate \Ours{} on 61 real-robot greenhouse trials per method (366 executions
in total). \Ours{}
achieves $\mathbf{75.41\%}$ average task success and $\mathbf{8.20\%}$ average
obstacle collision rate, outperforming representative imitation-learning
baselines and improving generalization in cluttered agricultural scenes.
\end{abstract}

\keywords{Imitation learning, Diffusion policy, Obstacle-aware representations}

\section{Introduction}
\label{sec:intro}
Robot manipulation in cluttered, real-world environments has long been a
central challenge in robotics, with applications spanning household
assistance, logistics, and agricultural automation~\citep{bac2014harvesting,
magistri2024improving}. Unlike controlled industrial settings, such
environments are characterized by two intertwined difficulties.
First, the robot must interact with diverse target objects while navigating
through unstructured obstacles---wires, tools, foliage, or stems---that form
narrow and irregular passages around the target.
Second, the surrounding scene exhibits substantial visual variability across
viewpoints, illumination conditions, and temporal changes, producing
similar-yet-non-identical observations that are difficult to handle with
fixed perception
pipelines~\citep{lobefaro2025spatio, sivakumar2021learned, yi2024view}.
Agricultural scenarios such as greenhouse manipulation represent a
particularly demanding instance of this setting, where dense plant
structures, continuous growth, and diurnal illumination changes amplify both
challenges simultaneously.
Traditional approaches typically rely on explicit, expert-designed
pipelines that encode task-specific knowledge through carefully engineered
perception and planning modules~\citep{magistri2024improving, lehnert2017autonomous,
wang2025ehc,wang2023optimizing}.
Such methods require extensive manual calibration and parameter tuning,
and their adaptability is inherently limited when deployed in environments
with significant clutter and visual variation.

Learning-based methods, particularly imitation learning (IL), have demonstrated promising performance in robotic manipulation tasks~\citep{zhao2023act,chi2023diffusion}. However, extending IL to cluttered, real-world scenes with unstructured obstacles surfaces several intertwined challenges. Unstructured obstacles form narrow, irregular passages around the target that IL methods trained in clean tabletop settings are ill-equipped to navigate, while visually complex backgrounds introduce distractors that further degrade perception-based policies~\citep{burns2024whatmakes,zhu2023groot,mirjalili2025arro}. Target appearance, illumination, and camera viewpoints also vary substantially across deployments, and policies trained on limited demonstrations frequently fail to generalize under such visual shifts~\citep{tobin2017domain,garcia2023sim2real}. Compounding these issues, standard end-to-end visual encoders entangle target, obstacle, and background cues into a single latent feature, leaving the downstream policy without an explicit signal about which parts of the scene must be \emph{reached} and which must be \emph{avoided}; as a result, current diffusion-based policies tend to produce trajectories that ignore nearby obstacles, especially under distribution shift~\citep{li2025lano3dp}.

To address these limitations, we present \textbf{\Ours{}}, a decomposed
diffusion-policy framework for imitation learning in cluttered, unstructured
scenes. \Ours{} uses an \emph{obstacle-aware visual encoder} to transform
RGB-D observations into a structured \emph{target--obstacle--background}
representation, making the target to reach and the obstacles to avoid explicit
to the downstream policy. Our main contributions are:

\begin{itemize}
  \item \textbf{Obstacle-aware structured representation.}
  We introduce a target--obstacle--background representation and a lightweight
  encoder that preserve the semantic roles of target, obstacle, and background
  rather than treating the input as a generic RGB-D image. This structure
  highlights target--obstacle spatial relationships and suppresses irrelevant
  background variation, providing more informative conditioning for
  diffusion-based policy learning.

  \item \textbf{Modular decomposed policy framework.}
  We decompose manipulation into diffusion-based alignment and replay-based
  interaction, decoupling obstacle-aware approach from task-specific close-range
  behavior. This allows the same learned alignment policy to support different
  interaction objectives by changing only the replay library, with a
  data-driven bottleneck rule selecting the interaction trajectory.

  \item \textbf{Real-world validation.}
  We evaluate \Ours{} on 61 real-robot greenhouse trials per method (366 executions in total), spanning target-pose,
  obstacle-layout, and target-appearance generalization. \Ours{} achieves a
  $75.41\%$ average task success rate and reduces the average obstacle collision
  rate to $8.20\%$, outperforming representative imitation-learning baselines
  and demonstrating improved generalization in cluttered agricultural scenes.
\end{itemize}

\section{Related Work}
\label{sec:related}

\subsection{Visual Imitation Learning and Representations}
Visual imitation learning has progressed from spatial action prediction and multimodal behavior cloning \cite{zeng2021transporter,florence2022implicit} to temporally coherent policies such as ACT \cite{zhao2023act}. Recent work shows that representation quality is critical for robust manipulation, including visual pretraining \cite{nair2022r3m,radosavovic2023real,burns2024whatmakes}, object-centric 3D features \cite{zhu2023groot}, and compact point-cloud policies such as DP3 \cite{ze2024dp3}. \Ours{} differs by using a task-structured target--obstacle--background representation instead of a generic image or point-cloud backbone, making the spatial relation between the target and obstacles explicit.

\subsection{Diffusion Policies for Manipulation}
Diffusion models are effective generative models for multimodal distributions \cite{ho2020ddpm,song2021ddim} and have been adapted to robotic planning and control through trajectory denoising and receding-horizon action generation \cite{janner2022diffuser,chi2023diffusion}. Subsequent diffusion-policy methods improve generalization with 3D conditioning \cite{ze2024dp3,ke2025diffuseractor}, equivariance \cite{wang2025equivariant,yang2025equibot}, or reusable sparse policy structure \cite{wang2025sparsedp}. These works focus mainly on stronger action generation or generic scene representations, whereas \Ours{} explicitly conditions diffusion-policy alignment on obstacle-aware visual structure.

\subsection{Obstacle-Aware Manipulation in Cluttered Scenes}
Cluttered agricultural scenes require reaching targets through foliage, stems, and narrow passages, and prior systems often rely on engineered perception, navigation, view planning, or shape completion \cite{bac2014harvesting,magistri2024improving,sivakumar2021learned,yi2024view,lehnert2017autonomous}. Learning-based motion methods can generate obstacle-aware trajectories from data \cite{qureshi2021mpnet,fishman2023mpinets,carvalho2023mpd}, but are typically planners rather than closed-loop visuomotor imitation policies. In contrast, strong imitation baselines such as ACT, Diffusion Policy, and DP3 \cite{zhao2023act,chi2023diffusion,ze2024dp3} do not explicitly separate regions to reach from regions to avoid. \Ours{} bridges this gap with a compact target--obstacle--background representation for obstacle-aware alignment before close-range interaction.


\section{Methodology}
\label{sec:method}

\begin{figure}[thbp]
    \centering
    \vspace{-0.1in}
    \includegraphics[width=\linewidth]{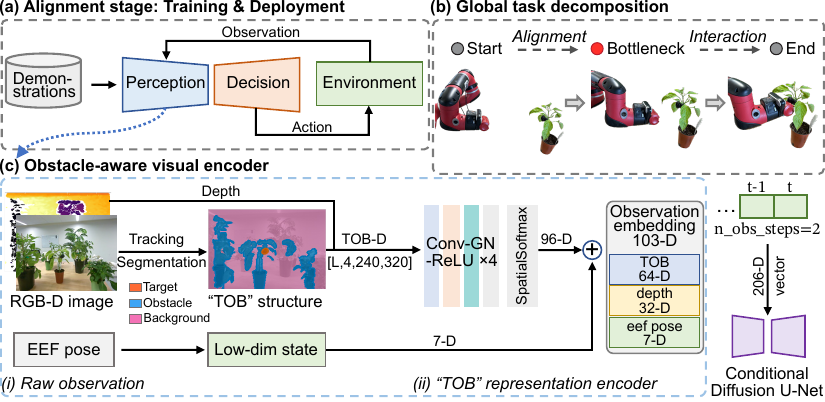}
    \caption{
\textbf{Overview of \Ours{}.}
Eye-in-hand RGB-D observations are converted into a structured
\emph{target--obstacle--background} (TOB) representation and encoded with robot
state into a 103-D conditioning vector for a diffusion policy. The policy
performs obstacle-aware alignment to a target-centered bottleneck pose, after
which replayed actions execute close-range interaction.
    }
    \vspace{-0.15in}
    \label{fig:overview}
\end{figure}

As illustrated in Figure~\ref{fig:overview}, \Ours{} is built around two design
choices for generalizable imitation learning in cluttered greenhouse scenes:
a structured visual encoder that exposes target--obstacle--background semantics
to the policy, and a decomposed execution scheme that separates visually guided
alignment from close-range interaction.

\subsection{Problem Formulation}
\label{sec:problem}

We formulate greenhouse manipulation as a two-stage visuomotor policy. At each
time step $t$, the robot observes
\begin{equation}
    \mathbf{o}_t = \{\mathbf{x}_t, \mathbf{e}_t\},
\end{equation}
where $\mathbf{x}_t \in \mathbb{R}^{4 \times H_o \times W_o}$ denotes the
structured visual observation and $\mathbf{e}_t \in \mathbb{R}^{7}$ denotes the
current end-effector pose. The resolution $H_o \times W_o$ corresponds to the
preprocessed policy input; in our implementation, $H_o=240$ and $W_o=320$. Both
$\mathbf{e}_t$ and the action $\mathbf{a}_t$ are represented as 7-D Cartesian
poses, consisting of a 3-D position and a unit quaternion.

We decompose the task at a target-centered bottleneck pose
$\mathbf{p}^{*} \in \mathbb{R}^{7}$:
\begin{equation}
  \pi(\mathbf{a}_{0:T} \mid \mathbf{o}_{0:T})
  =
  \underbrace{
  \pi_{\mathrm{align}}(\mathbf{a}_{0:T_b} \mid \mathbf{o}_{0:T_b})
  }_{\text{alignment stage}}
  \cdot
  \underbrace{
  \pi_{\mathrm{inter}}(\mathbf{a}_{T_b:T} \mid \mathbf{p}^{*})
  }_{\text{interaction stage}},
  \label{eq:decompose}
\end{equation}
where $T_b$ is not pre-specified but determined online as the first alignment
step at which the gated nearest-neighbor criterion of Sec.~\ref{sec:bottleneck_replay}
is satisfied, after which control switches to trajectory replay.

\subsection{Structured Target--Obstacle--Background Observation}
\label{sec:tob}

\Ours{} replaces raw RGB-D inputs with a structured
\emph{target--obstacle--background} (TOB) observation that exposes the scene
semantics most relevant to obstacle-aware alignment. Given an eye-in-hand RGB-D frame, a fine-tuned YOLOv8s-Worldv2~\citep{cheng2024yolo} open-vocabulary detector
predicts the target and obstacle regions. Pixels assigned to the target region are labeled as target, pixels assigned to obstacle regions are labeled as obstacle, and all remaining pixels are labeled as background. This produces a three-class semantic index map, which is converted into three binary masks and concatenated with the aligned, normalized depth image:
\begin{equation}
    \mathbf{x}_t =
    \left[
    \mathbf{m}^{\mathrm{bg}}_t;\,
    \mathbf{m}^{\mathrm{obs}}_t;\,
    \mathbf{m}^{\mathrm{tar}}_t;\,
    \bar{\mathbf{d}}_t
    \right]
    \in [0,1]^{4 \times H_o \times W_o},
    \label{eq:tob}
\end{equation}
where $\mathbf{m}^{\mathrm{bg}}_t$, $\mathbf{m}^{\mathrm{obs}}_t$, and
$\mathbf{m}^{\mathrm{tar}}_t$ are the background, obstacle, and target masks,
respectively, and $\bar{\mathbf{d}}_t$ is the normalized depth image. Details of
RGB-D preprocessing and YOLO-based mask construction are provided in
Appendix~\ref{app:tob_details}.

By encoding semantic roles and depth in a compact four-channel input, the TOB
observation preserves target--obstacle spatial relationships while suppressing
appearance-level background variation. This provides the diffusion policy with
high-level structure for generating obstacle-aware approach motions.

\subsection{Obstacle-aware Structured Observation Encoder}
\label{sec:encoder}

The TOB observation is not a natural image, but a structured representation
whose channels have explicit semantic roles: background, obstacle, target, and
depth. Therefore, we do not process it with a generic image backbone such as the
ResNet18 used in Diffusion Policy~\cite{chi2023diffusion}. Treating TOB as
a standard image would ignore its channel-wise structure and force the encoder
to relearn spatial abstractions that are already exposed by the preprocessing
stage. Instead, we use a lightweight \emph{StructuredObsEncoder} that preserves
the separation between semantic masks and depth and extracts compact geometric
features for policy conditioning.

The encoder splits the TOB observation into a semantic branch and a depth
branch:
\begin{equation}
    \mathbf{x}^{\mathrm{seg}}_t = \mathbf{x}_t[0:3,:,:],
    \qquad
    \mathbf{x}^{\mathrm{dep}}_t = \mathbf{x}_t[3,:,:].
\end{equation}
The semantic branch encodes the background, obstacle, and target masks, while
the depth branch encodes normalized geometric information. Each branch uses a
small convolutional network followed by spatial softmax, which converts feature
maps into keypoint-like coordinates rather than dense image features. This
design encourages the encoder to capture compact spatial relations, especially
the relative layout between the target and nearby obstacles.

The semantic and depth features are concatenated with the end-effector pose:
\begin{equation}
    \mathbf{c}_t =
    \left[
    \mathbf{z}^{\mathrm{seg}}_t;\,
    \mathbf{z}^{\mathrm{dep}}_t;\,
    \mathbf{e}_t
    \right]
    \in \mathbb{R}^{103},
    \label{eq:encoder_output}
\end{equation}
where $\mathbf{z}^{\mathrm{seg}}_t \in \mathbb{R}^{64}$,
$\mathbf{z}^{\mathrm{dep}}_t \in \mathbb{R}^{32}$, and
$\mathbf{e}_t \in \mathbb{R}^{7}$. With two observation steps, the diffusion
policy receives a $206$-D conditioning vector. Architectural details are
provided in Appendix~\ref{app:encoder_details}.

\subsection{Alignment-stage Diffusion Policy}
\label{sec:diffusion}

The alignment policy is implemented as a conditional diffusion policy that
predicts a sequence of end-effector actions for reaching the bottleneck pose.
Let
\begin{equation}
    \mathbf{A}^{0}
    =
    \{\mathbf{a}_1,\ldots,\mathbf{a}_{T_a}\}
\end{equation}
denote a clean action sequence from the alignment stage. During training, we
corrupt $\mathbf{A}^{0}$ with Gaussian noise through a forward diffusion
process:
\begin{equation}
  q(\mathbf{A}^{k} \mid \mathbf{A}^{0})
  =
  \mathcal{N}
  \left(
  \mathbf{A}^{k};
  \sqrt{\bar{\alpha}_k}\mathbf{A}^{0},
  (1-\bar{\alpha}_k)\mathbf{I}
  \right),
  \label{eq:forward}
\end{equation}
where $k$ is the diffusion step and
$\bar{\alpha}_k=\prod_{i=1}^{k}(1-\beta_i)$.

A noise-prediction network $\epsilon_{\theta}$ is trained to denoise the action
sequence conditioned on the encoded structured observations:
\begin{equation}
  \mathcal{L}
  =
  \mathbb{E}_{k,\mathbf{A}^{0},\boldsymbol{\epsilon}}
  \left[
  \left\|
  \boldsymbol{\epsilon}
  -
  \epsilon_{\theta}
  \left(
  \mathbf{A}^{k}, \mathbf{c}_{1:n}, k
  \right)
  \right\|_2^2
  \right],
  \label{eq:loss}
\end{equation}
where $\boldsymbol{\epsilon}\sim\mathcal{N}(\mathbf{0},\mathbf{I})$ and
$\mathbf{c}_{1:n}$ denotes the conditioning features from the most recent
$n$ observation steps. In our experiments, $n=2$.

At inference time, the policy samples an action sequence by iterative denoising:
\begin{equation}
  \mathbf{A}^{k-1}
  =
  \frac{1}{\sqrt{\alpha_k}}
  \left(
  \mathbf{A}^{k}
  -
  \frac{\beta_k}{\sqrt{1-\bar{\alpha}_k}}
  \epsilon_{\theta}(\mathbf{A}^{k},\mathbf{c}_{1:n},k)
  \right)
  +
  \sigma_k \mathbf{z},
  \label{eq:reverse}
\end{equation}
where $\mathbf{z}\sim\mathcal{N}(\mathbf{0},\mathbf{I})$. The first 8 actions from the predicted sequence of 16 are executed in a receding horizon manner before re-planning. Since the
conditioning vector explicitly encodes target, obstacle, background, depth, and
end-effector pose, the diffusion policy is encouraged to generate approach
motions that move toward the target while avoiding nearby plant obstacles.

\subsection{Bottleneck Switching and Interaction}
\label{sec:bottleneck_replay}

This modular design decouples obstacle-aware approach from task-specific
close-range behavior, allowing the same learned alignment policy to support
different interaction objectives by changing only the replay library. For
example, the interaction stage can be adapted for touching, harvesting, or
close-up inspection without retraining the diffusion policy. We therefore use a
data-driven bottleneck switching criterion to transition from the learned
alignment phase to the replay-based interaction phase.

The alignment policy is trained on trajectory segments before contact-rich
interaction and drives the robot toward a configuration from which an
interaction trajectory can be reliably replayed. Rather than switching at a
manually hard-coded pose, \Ours{} compares the current target observation with
the initial target observations stored in a replay library collected from
demonstrations. The replay library contains $N_{\mathrm{lib}}$ interaction
trajectories:
\begin{equation}
    \mathcal{D}_{\mathrm{replay}}
    =
    \left\{
    \left(\mathbf{g}^{i}_{0}, \mathcal{A}^{i}_{\mathrm{inter}}\right)
    \right\}_{i=1}^{N_{\mathrm{lib}}},
\end{equation}
where $\mathbf{g}^{i}_{0}$ denotes the target state at the first frame of the
$i$-th interaction trajectory, and $\mathcal{A}^{i}_{\mathrm{inter}}$ denotes
the corresponding replay action sequence. During deployment, if the current
target state is sufficiently close to an entry in the replay library, the
system switches to interaction and replays the nearest corresponding action
sequence. This lightweight transition mechanism also avoids relying on
eye-in-hand depth during close-range interaction, where depth observations can
be unreliable. Implementation details are provided in
Appendix~\ref{app:bottleneck_replay_details}.

\section{Experiments}
\label{sec:experiments}

\subsection{Experimental Protocol}
\label{sec:protocol}

\textbf{Hardware.}
All real-robot experiments were conducted on a 7-DoF Sawyer arm with an
eye-in-hand Intel RealSense D455 RGB-D camera, used directly as the onboard
observation sensor without external camera calibration. The policy was deployed
on an NVIDIA RTX 2080 Ti workstation and trained on an NVIDIA RTX 6000 Ada
workstation.

\begin{wrapfigure}{t}{0.5\textwidth}
\vspace{-0.17in}
  \centering
  \includegraphics[
    width=\linewidth,
  ]{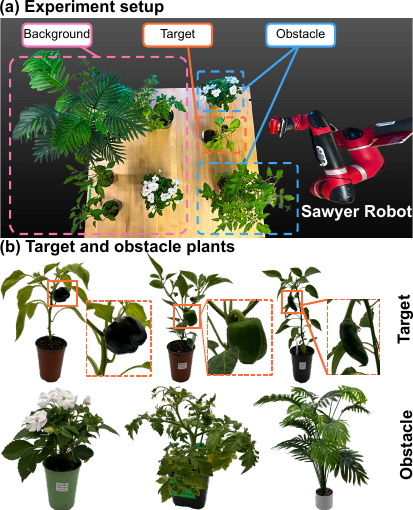}
\caption{\textbf{Experimental setup.}
(a) Indoor greenhouse testbed with a Sawyer robot, target pepper, background plants, and obstacle plants.
(b) Target pepper plants and obstacle plants used in the experiments.
}
  \label{fig:setup}
  \vspace{-0.15in}
\end{wrapfigure}

\textbf{Task Benchmarks.}
We evaluate \Ours{} in an indoor greenhouse testbed that reproduces the visual
clutter and spatial constraints of real greenhouse manipulation. The task
requires the robot to reach a pepper target within a predefined target region
while avoiding surrounding plant obstacles, as shown in Fig.~\ref{fig:setup}.
We design three evaluation settings to test generalization along different
axes.

First, \emph{Target-Pose Generalization} (\textbf{TPG}) evaluates robustness to
changes in the target pose. We select 10 target positions within the target
region and use two target orientations, resulting in 20 trials. Second,
\emph{Obstacle-Layout Generalization} (\textbf{OLG}) keeps the target fixed and
varies the surrounding obstacle plants. We use three obstacle orderings and
seven obstacle positions, resulting in 21 trials. Third, \emph{Target-Appearance
Generalization} (\textbf{TAG}) changes the target from the training purple
pepper to two unseen green pepper varieties. For each green pepper type, we
place the target in 10 different poses within the target region, resulting in
20 trials. Together, TPG, OLG, and TAG comprise 61 evaluation trials and test
whether the policy can generate obstacle-aware approach motions under target
pose shifts, obstacle rearrangements, and unseen target appearances.
Each trial is a separate real-robot execution under a distinct configuration
rather than a repetition of a fixed configuration, so the protocol measures
configuration-level generalization rather than within-configuration
repeatability. All 61 trials are run \emph{per method}, giving $61 \times 6 = 366$
real-robot executions in total. For OLG, training and evaluation use the same
three obstacle plant species, but the evaluated spatial arrangements and
positions are not seen during demonstration collection; OLG therefore measures
generalization to unseen layouts and positions of known obstacle plants, not to
unseen obstacle categories.

\textbf{Demonstrations.}
We collected 90 expert alignment demonstrations using the gravity-compensation
teaching mode of the Sawyer robot. All demonstrations use the purple pepper
target, making the green pepper targets in TAG fully unseen during training.
The demonstrations are evenly distributed across three obstacle configurations,
with 30 target-pose demonstrations per configuration. Each trajectory records
the motion from the home position to the bottleneck pose.
The interaction stage is handled separately by a replay library containing 26 trajectories from the bottleneck pose to task completion, defined as touching
the target pepper. At deployment, \Ours{} selects the closest replay trajectory
according to the bottleneck switching rule in Sec.~\ref{sec:bottleneck_replay}.

\textbf{Training Details.}
All diffusion-policy variants (\DPrgb{}, \DPrgbd{}, \DPtob{}, and \Ours{}) share
an identical pipeline: the same 90 demonstrations, the same train/validation
split (random seed 42, $10\%$ held out), two observation steps, prediction
horizon 16, action horizon 8, and 250 epochs with AdamW ($1\times10^{-4}$
learning rate, $1\times10^{-6}$ weight decay, betas $(0.95,0.999)$). We used EMA
evaluation and a DDIM scheduler with squared-cosine noise, $\epsilon$-prediction,
and $K=100$ diffusion steps for both training and inference. For \DPrgbd{} and
\DPtob{}, only the first convolution of the ResNet-18 encoder is expanded from
three to four input channels; every remaining layer is unchanged. The sole
remaining discrepancy among the DP variants is the per-device batch size (32 for
\DPtob{} and \Ours{}, 16 for \DPrgb{} and \DPrgbd{}). Because \DPtob{} and
\Ours{} share the same batch size, batch size does not confound the comparison
that isolates the visual encoder.

ACT baselines used chunk size 16, batch size 8, hidden dimension 512, feedforward
dimension 3200, KL weight 10, and learning rate $1\times10^{-5}$, and were trained
for 2000 epochs. All methods were evaluated using the checkpoint with the lowest
validation loss, with early stopping.

\textbf{Baselines.}
We compare \Ours{} with representative imitation-learning baselines and input
ablations. All methods use the same 90 demonstrations, the same bottleneck
switching rule, and the same 26-trajectory replay library, so the decomposition
cannot explain the differences reported below; what varies is the learned
alignment policy and its observation representation.

\begin{itemize}
  \item \textbf{\ACTrgb}~\citep{zhao2023act}: an action chunking transformer policy
  trained with standard RGB observations for the alignment stage.

  \item \textbf{\ACTtob}: ACT conditioned on the proposed
  target--obstacle--background observation, used to evaluate whether TOB also
  benefits non-diffusion imitation policies.

  \item \textbf{\DPrgb}~\citep{chi2023diffusion}: the standard Diffusion Policy
  with a ResNet-18 visual encoder over RGB observations.

  \item \textbf{\DPrgbd}: the same Diffusion Policy encoder with the aligned
  depth image appended as a fourth input channel, used to isolate the
  effect of adding depth alone.

  \item \textbf{\DPtob}: the same Diffusion Policy encoder taking exactly the
  four-channel TOB observation used by \Ours{}, used to isolate the
  effect of the observation representation from that of the structured encoder.

  \item \textbf{\Ours}: the full method using TOB observations, the
  structured observation encoder, and diffusion-based alignment.
\end{itemize}

\textbf{On point-cloud baselines.}
We do not report DP3~\citep{ze2024dp3} as a quantitative baseline. DP3 conditions
on sampled 3D point clouds reconstructed from a fixed external camera, whereas
our setting uses image-aligned observations from an eye-in-hand D455 without
constructing explicit 3D geometry. Because the camera frame moves continuously
with the arm, and because thin foliage, occlusion boundaries, and missing depth
introduce substantial reconstruction artifacts, our implementation produced
unstable point clouds under registration, cropping, and sampling, and we were
unable to deploy it reliably. We therefore report this practical mismatch rather
than statistics from an under-tuned baseline. This reflects the difficulty of
transferring DP3 to a moving eye-in-hand setup in dense foliage, not an inherent
restriction of the method to fixed-camera settings.

\textbf{Evaluation Metrics.}
We evaluate each method on three generalization settings: Target-Pose
Generalization (TPG), Obstacle-Layout Generalization (OLG), and
Target-Appearance Generalization (TAG). We report task success rate (TSR) as
the primary metric, defined as the percentage of trials in which the robot
reaches and interacts with the target without safety violations. We also report
obstacle collision rate (OCR), task completion time (TCT), and trajectory smoothness
(Traj. Smooth). OCR is the percentage of
trials with obstacle collisions, TCT is the wall-clock time from motion onset to task completion, averaged over
successful TPG trials, and
trajectory smoothness measures the temporal variation of end-effector poses and actions. Higher TSR is better; lower OCR,
TCT, and smoothness values are better.

\subsection{Main Results}
\label{sec:main}

\begin{table*}[ht]
\centering
\caption{
Main real-robot results across three generalization settings. Each method is run
on all 61 trials (TPG $n{=}20$, OLG $n{=}21$, TAG $n{=}20$), giving 366 real-robot
executions in total; the mean is computed over the 61 pooled trials rather than
as an average of the three per-setting rates. TCT and trajectory smoothness are
computed over successful TPG trials. Best results are shown in bold.
}
\label{tab:main_results}
\resizebox{\textwidth}{!}{
\begin{tabular}{lcccccccccc}
\toprule
\multirow{2}{*}{\textbf{Method}}
& \multicolumn{4}{c}{\textbf{TSR (\%) $\uparrow$}}
& \multicolumn{4}{c}{\textbf{OCR (\%) $\downarrow$}}
& \multirow{2}{*}{\textbf{TCT (s) $\downarrow$}}
& \multirow{2}{*}{\textbf{Smooth ($10^{-4}$)$ \downarrow$}} \\

\cmidrule(lr){2-5}
\cmidrule(lr){6-9}
& \textbf{TPG}
& \textbf{OLG}
& \textbf{TAG}
& \textbf{Mean}
& \textbf{TPG}
& \textbf{OLG}
& \textbf{TAG}
& \textbf{Mean}
&
& \\
\midrule
\ACTrgb
& 40.00 & 52.38 & 45.00 & 45.90
& 25.00 & \textbf{14.29} & 5.00 & 14.75
& 20.77 & 1.16 \\
\ACTtob
& 30.00 & 66.67 & 25.00 & 40.98
& 15.00 & \textbf{14.29} & 5.00 & 11.48
& 18.38 & \textbf{0.77} \\
\midrule
\DPrgb
& 55.00 & 42.86 & 50.00 & 49.18
& 35.00 & 33.33 & 10.00 & 26.23
& 18.41 & 1.35 \\
\DPrgbd
& 60.00 & 42.86 & 50.00 & 50.82
& 35.00 & 33.33 & 10.00 & 26.23
& 18.67 & 1.34 \\
\DPtob
& 55.00 & 38.10 & 45.00 & 45.90
& 30.00 & 38.10 & 15.00 & 27.87
& 18.17 & 1.19 \\
\midrule
\textbf{\Ours}
& \textbf{75.00} & \textbf{71.43} & \textbf{80.00} & \textbf{75.41}
& \textbf{5.00} & 19.05 & \textbf{0.00} & \textbf{8.20}
& \textbf{15.57} & 1.13 \\
\bottomrule
\end{tabular}
}
\end{table*}

Fig.~\ref{fig:teaser}(c) and Table~\ref{tab:main_results} summarize the main
real-robot results. \Ours{} achieves the best task success rate in all three
generalization settings, with a mean TSR of $\textbf{75.41\%}$ against $45.90\%$
for \ACTrgb{} and $49.18\%$ for \DPrgb{}, and the best mean OCR
($\textbf{8.20\%}$ versus $14.75\%$ and $26.23\%$). Neither of the two
information-matched diffusion controls closes this gap: \DPrgbd{} reaches
$50.82\%$ TSR and \DPtob{} $45.90\%$, i.e. $31/61$ and $28/61$ successes against
$30/61$ for \DPrgb{}. We analyze these controls in Sec.~\ref{sec:ablation}. The improvement is consistent under
target-pose shifts (TPG), obstacle rearrangements (OLG), and unseen target
appearances (TAG), including the unseen green pepper targets shown in
Fig.~\ref{fig:setup}(b). These results indicate that the TOB representation improves not only task
completion but also the reliability of obstacle-aware approach motions in
cluttered greenhouse scenes.

\textbf{Qualitative Analysis.}
Qualitatively, baseline policies often move directly toward the target and are
more likely to contact nearby plants in cluttered layouts. In contrast,
\Ours{} produces more obstacle-aware approach motions by aligning with the
target-centered free space before entering the bottleneck region. This behavior
is consistent with the lower average OCR in Table~\ref{tab:main_results},
indicating that the TOB representation provides useful spatial cues for
distinguishing the target from nearby obstacles.

\subsection{Ablation Studies}
\label{sec:ablation}

\begin{wrapfigure}{t}{0.5\textwidth}
\vspace{-0.1in}
  \centering
  \includegraphics[
    width=\linewidth,
  ]{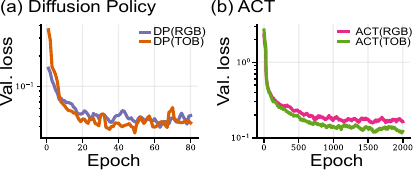}

\caption{
\textbf{TOB stabilizes training.}
TOB reduces validation loss, but the \ACTtob{}
real-robot results suggest that lower supervised loss alone does not guarantee higher task success.
}
  \label{fig:ablation}

  \vspace{-0.1in}
\end{wrapfigure}

We first isolate the contribution of the TOB representation from that of the
structured encoder using the two information-matched diffusion controls in
Table~\ref{tab:main_results}. Appending aligned depth to the standard DP encoder
moves mean TSR only from $49.18\%$ to $50.82\%$---one additional successful trial
out of 61---and leaves OCR unchanged at $26.23\%$; the collision counts of
\DPrgb{} and \DPrgbd{} coincide in every setting ($7/20$, $7/21$, $2/20$),
indicating that depth alone contributes little to obstacle avoidance under this
encoder. Feeding the same encoder the identical four-channel TOB observation
used by \Ours{} does not help either: mean TSR falls to $45.90\%$ and mean OCR
rises to $27.87\%$. The degradation is concentrated precisely in the setting TOB
was designed to address---under obstacle-layout shift, \DPtob{} records the
lowest success rate and the highest collision rate of any method in
Table~\ref{tab:main_results} (both $8/21$, i.e. $38.10\%$). Since \DPtob{} and
\Ours{} share the same observation, the same demonstrations, and the same batch
size, the remaining $29.5$-point TSR gap is attributable to the structured
encoder. Structured semantic input alone therefore does not confer obstacle
awareness; without an encoder matched to its channel structure it is no better
than raw RGB, and can be worse.

For the non-diffusion backbone, adding TOB to ACT reduces mean OCR from
$14.75\%$ to $11.48\%$ and improves trajectory smoothness from $1.16$ to $0.77$,
but mean TSR drops from $45.90\%$ to $40.98\%$---the same pattern, on a different
policy class. We do not claim the convolutional backbone or the spatial softmax
as novel components; the contribution is their task-specific coupling with TOB.
The validation curves in Fig.~\ref{fig:ablation} are consistent with this view:
TOB reduces validation-loss fluctuation, but lower supervised loss alone does not
translate into higher task success.

\subsection{Discussion}
\Ours{} targets visual imitation learning in cluttered real-world scenes,
where small targets are surrounded by leaves, stems, and background distractors.
The target--obstacle--background (TOB) representation makes these semantic roles
explicit, helping the diffusion policy focus on the target to reach and the
obstacles to avoid rather than raw visual clutter. Its decomposed design further
supports practical eye-in-hand deployment: RGB-D observations guide the approach
stage, while replayed actions handle close-range interaction where depth can
become unreliable.

\section{Limitations}
\label{sec:limitations}

\Ours{} currently relies on human- or prompt-specified targets, which lets us
focus on low-level obstacle-aware alignment but does not address autonomous
target selection or task planning. It also inherits the common limitation of
imitation learning: the policy is constrained by the demonstration distribution
and has limited ability to explore behaviors outside the collected data. 
Beyond these, only the alignment stage is learned; the interaction stage replays
demonstrated trajectories, so our results should be read as evidence about
obstacle-aware approach behavior rather than about contact-rich manipulation.
Our evaluation is also confined to a single indoor greenhouse testbed with three
obstacle plant species and one target crop, and the comparison against
point-cloud policies remains open in the moving eye-in-hand regime.
\section{Conclusion}
\label{sec:conclusion}

This work presented \textbf{\Ours{}}, a diffusion-policy framework for
imitation learning in cluttered and unstructured greenhouse environments. The
central contribution is a lightweight obstacle-aware encoder that converts
eye-in-hand RGB-D observations into a structured
\emph{target--obstacle--background} (TOB) representation. By separating the
target to reach, the obstacles to avoid, and the background to ignore, TOB
provides the diffusion policy with high-level spatial structure for
obstacle-aware action generation and improves generalization beyond raw visual
imitation.

\Ours{} also uses a deployment-oriented decomposed execution scheme, where the
learned policy handles visually guided alignment and replayed actions handle
close-range interaction. This design avoids relying on unreliable eye-in-hand
depth during contact-rich motions and allows different interaction objectives to
be supported by changing the replay library. Real-robot experiments under
target-pose, obstacle-layout, and target-appearance shifts show that \Ours{}
improves task success and obstacle avoidance over imitation-learning baselines.
Future work will extend \Ours{} to mobile manipulation and integrate
vision-language models for autonomous target selection.

\acknowledgments{We gratefully acknowledge support from the United States
Department of Agriculture (Grant No.~2024-67021-42528) and the National Science
Foundation (Grant No.~2551220).}

\bibliography{ref}

\clearpage
\appendix

\section{Implementation details}

\subsection{Target--Obstacle--Background Preprocessing Details}
\label{app:tob_details}

This section describes the preprocessing pipeline used to construct the
target--obstacle--background (TOB) observation.

\paragraph{RGB-D acquisition.}
The RGB image is obtained from the eye-in-hand camera stream
\texttt{/camera/color/image\_raw} as
$\mathbf{I}_t \in \mathbb{R}^{H \times W \times 3}$ with \texttt{uint8}
values. The depth image is obtained from
\texttt{/camera/aligned\_depth\_to\_color/image\_raw} as
$\mathbf{D}_t \in \mathbb{R}^{H \times W}$ with \texttt{uint16} values in
millimeters. The depth image is pixel-aligned to the RGB image, and zero depth
values denote invalid pixels. Invalid depth values are temporally hole-filled
in the depth callback.

\paragraph{Infrared projector artifact reduction.}
To reduce artifacts from the RGB-D camera infrared projector, we use an
emitter even--odd alternation procedure during acquisition. This produces a
cleaner RGB stream for downstream YOLO parsing.
Depth frames are saved on even frames with the emitter enabled. The RGB and depth buffers are then truncated to the same length before saving.

\paragraph{Semantic parsing.}
We use YOLOv8s-Worldv2~\citep{cheng2024yolo} as the open-vocabulary
detector, fine-tuned on approximately 80 manually annotated greenhouse images
collected from the demonstration sessions. The target class denotes the selected
manipulation target part, namely the pepper fruit to be touched, rather than the
whole plant. Detected obstacle boxes are refined into pixel-accurate masks with
MobileSAM~\citep{mobile_sam}. The resulting per-pixel labels form the
semantic index map $\mathbf{s}_t \in \{0,1,2\}^{H \times W}$, where label 0
denotes background, label 1 denotes obstacle, and label 2 denotes the target;
pixels assigned to neither the target nor an obstacle are treated as background.
The index map is converted into three binary masks:
\begin{align}
    \mathbf{m}^{\mathrm{bg}}_t(u,v)
    &= \mathbbm{1}\!\left[\mathbf{s}_t(u,v)=0\right], \\
    \mathbf{m}^{\mathrm{obs}}_t(u,v)
    &= \mathbbm{1}\!\left[\mathbf{s}_t(u,v)=1\right], \\
    \mathbf{m}^{\mathrm{tar}}_t(u,v)
    &= \mathbbm{1}\!\left[\mathbf{s}_t(u,v)=2\right].
\end{align}

\paragraph{Four-channel policy input.}
The three semantic masks and the aligned depth image are resized to the policy
input resolution $H_o \times W_o = 240 \times 320$. The depth image is
normalized to $[0,1]$:
\begin{equation}
    \bar{\mathbf{d}}_t = \mathrm{Normalize}(\mathbf{D}_t),
\end{equation}
where invalid pixels are set to 0. The
final policy input is
\begin{equation}
    \mathbf{x}_t =
    \left[
    \mathbf{m}^{\mathrm{bg}}_t;\,
    \mathbf{m}^{\mathrm{obs}}_t;\,
    \mathbf{m}^{\mathrm{tar}}_t;\,
    \bar{\mathbf{d}}_t
    \right]
    \in [0,1]^{4 \times 240 \times 320}.
\end{equation}
In implementation, this tensor corresponds to \texttt{cam4}, whose channel
order is
\[
    [\texttt{bg\_mask},\ \texttt{obst\_mask},\ \texttt{plant\_mask},\
    \texttt{depth\_norm}].
\]
For batched policy training, the observation tensor is represented as
$\mathbf{X} \in \mathbb{R}^{B \times T_o \times 4 \times 240 \times 320}$,
where $B$ is the batch size and $T_o$ is the number of observation steps. In our
experiments, $T_o=2$.

\subsection{Structured Observation Encoder}
\label{app:encoder_details}

This section provides implementation details of the
\emph{StructuredObsEncoder} used to encode the four-channel TOB observation.

\paragraph{Input split.}
Given the TOB observation
$\mathbf{x}_t \in [0,1]^{4 \times 240 \times 320}$, the encoder splits the
input into a semantic tensor and a depth tensor:
\begin{equation}
    \mathbf{x}^{\mathrm{seg}}_t =
    \mathbf{x}_t[0:3,:,:]
    \in [0,1]^{3 \times 240 \times 320},
    \qquad
    \mathbf{x}^{\mathrm{dep}}_t =
    \mathbf{x}_t[3,:,:]
    \in [0,1]^{1 \times 240 \times 320}.
\end{equation}
The first three channels correspond to the background, obstacle, and target
masks, while the fourth channel corresponds to normalized depth.

\paragraph{Branch encoders.}
The semantic and depth branches use the same lightweight CNN structure but
different channel widths. Each branch consists of four convolutional blocks,
where each block is
\[
    \mathrm{Conv2D} \rightarrow \mathrm{GroupNorm} \rightarrow \mathrm{ReLU}.
\]
The convolutional layers use kernels and strides
\[
    (5{\times}5, s{=}2),\quad
    (3{\times}3, s{=}2),\quad
    (3{\times}3, s{=}1),\quad
    (3{\times}3, s{=}1).
\]
For a $240 \times 320$ input, the two stride-$2$ layers reduce the spatial
resolution to $60 \times 80$. The semantic branch maps
\begin{equation}
    \mathbb{R}^{3 \times 240 \times 320}
    \rightarrow
    \mathbb{R}^{32 \times 60 \times 80},
\end{equation}
and the depth branch maps
\begin{equation}
    \mathbb{R}^{1 \times 240 \times 320}
    \rightarrow
    \mathbb{R}^{16 \times 60 \times 80}.
\end{equation}

\paragraph{Spatial softmax.}
Instead of flattening the feature maps, each branch applies spatial softmax to
convert feature activations into keypoint-like coordinates. For channel $c$ of
a feature map $\mathbf{F}$, spatial softmax computes
\begin{equation}
    \mathbf{k}_{c}
    =
    \sum_{u,v}
    \mathrm{softmax}(\mathbf{F}_{c})(u,v)
    \begin{bmatrix}
    u \\ v
    \end{bmatrix}.
    \label{eq:spatial_softmax}
\end{equation}
Thus, each output channel contributes one expected 2-D coordinate. The semantic
branch produces $32$ coordinate pairs, giving
$\mathbf{z}^{\mathrm{seg}}_t \in \mathbb{R}^{64}$, while the depth branch
produces $16$ coordinate pairs, giving
$\mathbf{z}^{\mathrm{dep}}_t \in \mathbb{R}^{32}$.

\paragraph{Final conditioning vector.}
The branch outputs are concatenated with the 7-D end-effector pose:
\begin{equation}
    \mathbf{c}_t =
    \left[
    \mathbf{z}^{\mathrm{seg}}_t;\,
    \mathbf{z}^{\mathrm{dep}}_t;\,
    \mathbf{e}_t
    \right]
    \in \mathbb{R}^{103}.
\end{equation}
Since the policy uses two observation steps, the final conditioning input to the
diffusion U-Net has dimension $2 \times 103 = 206$.

\subsection{Bottleneck Switching and Interaction Details}
\label{app:bottleneck_replay_details}

This section describes the data-driven switching rule used to transition from
the alignment stage to the interaction stage.

\paragraph{Replay library.}
From the demonstration dataset, we construct a replay library
$\mathcal{D}_{\mathrm{replay}}$ containing $N_{\mathrm{lib}}$ interaction
samples:
\begin{equation}
    \mathcal{D}_{\mathrm{replay}}
    =
    \left\{
    \left(\mathbf{g}^{i}_{0}, \mathcal{A}^{i}_{\mathrm{inter}}\right)
    \right\}_{i=1}^{N_{\mathrm{lib}}},
\end{equation}
where $\mathcal{A}^{i}_{\mathrm{inter}}
= \{\hat{\mathbf{a}}^{i}_{0},\ldots,\hat{\mathbf{a}}^{i}_{S_i}\}$ is the
interaction-stage action sequence from the $i$-th demonstration, and
$\mathbf{g}^{i}_{0}$ is the target state extracted from the first frame of that
interaction segment. In our implementation,
$N_{\mathrm{lib}}=26$.

\paragraph{Target state.}
The target state is represented by the target center and target depth:
\begin{equation}
    \mathbf{g}_t
    =
    \left[
    u^{\mathrm{tar}}_t,\;
    v^{\mathrm{tar}}_t,\;
    d^{\mathrm{tar}}_t
    \right]^{\top},
\end{equation}
where $(u^{\mathrm{tar}}_t, v^{\mathrm{tar}}_t)$ is the center of the tracked
target bounding box in image coordinates, and $d^{\mathrm{tar}}_t$ is defined in
the switching rule below. The stored state $\mathbf{g}^{i}_{0}$ is computed in
the same way from the first frame of the $i$-th replay action sequence.

\paragraph{Nearest-neighbor switching rule.}
At each alignment step, we compare the current target state against the initial
target states stored in the replay library. The target depth
$d^{\mathrm{tar}}_t$ is computed from the aligned depth image as the
$p$-th percentile of valid depth pixels inside a central crop of the target
bounding box. In our experiments, we use the central $40\%$ of the target box
and $p=20$, which biases the estimate toward the visible foreground surface of
the target.

For the $i$-th replay entry, let
$\mathbf{g}^{i}_{0}=[u^{i}_{0},v^{i}_{0},d^{i}_{0}]^\top$ denote the target
state recorded at the first frame of that interaction replay. We compute the
normalized image-plane errors
\begin{equation}
    e_u^{i}
    =
    \frac{
    \left|u^{\mathrm{tar}}_t - u^{i}_{0}\right|
    }{W},
    \qquad
    e_v^{i}
    =
    \frac{
    \left|v^{\mathrm{tar}}_t - v^{i}_{0}\right|
    }{H},
\end{equation}
and the one-sided normalized depth error
\begin{equation}
    e_d^{i}
    =
    \max\left(
    0,\,
    \frac{
    d^{\mathrm{tar}}_t - d^{i}_{0}
    }{D}
    \right),
\end{equation}
where $W$ and $H$ are the RGB image width and height, and $D$ is the depth
normalization scale in millimeters. The one-sided depth error treats the
current target as depth-compatible if it is already closer to the camera than
the replay reference.

Switching uses a gated nearest-neighbor rule. Each replay entry maintains an
entry-specific depth-ready flag:
\begin{equation}
    r_i(t)
    =
    r_i(t-1)
    \lor
    \left[
    e_d^{i}(t) \leq \epsilon_d
    \right].
\end{equation}
That is, a replay entry becomes depth-ready once the live target depth has
entered its acceptable depth range at least once. Among depth-ready entries, we
select the replay with the smallest image-plane error
\begin{equation}
    i^*
    =
    \arg\min_{i:r_i(t)=1}
    \left(
    e_u^{i} + e_v^{i}
    \right).
\end{equation}
The system switches to the interaction stage when the selected replay also
satisfies
\begin{equation}
    e_u^{i^*} + e_v^{i^*}
    \leq
    \epsilon_{uv}.
\end{equation}
It then executes the interaction action sequence associated with that replay:
\begin{equation}
    \pi_{\mathrm{inter}}
    =
    \mathrm{Replay}
    \left(
    \mathcal{A}^{i^*}_{\mathrm{inter}}
    \right).
\end{equation}

In our experiments, we use
$\epsilon_d=0.005$ with $D=1000$ mm, corresponding to a 5 mm depth gate, and
$\epsilon_{uv}=0.2$ for the normalized image-plane error. We use
$W=640$ and $H=480$ for the image-plane normalization.

\paragraph{Motivation.}
This rule implements the bottleneck transition without requiring a hard-coded
Cartesian pose. Instead, it switches when the current target-centered visual
configuration is close to one observed at the beginning of a demonstrated
interaction trajectory. This is useful for eye-in-hand deployment because the
alignment policy can use structured RGB-D observations while the target is still
observable, whereas the subsequent close-range interaction relies on replayed
actions when depth around the contact region may become noisy or invalid.

\section{TOB Preprocessing Visualization}
\label{app:tob_visualization}

\begin{figure}[h]
    \centering
    \includegraphics[width=0.75\linewidth]{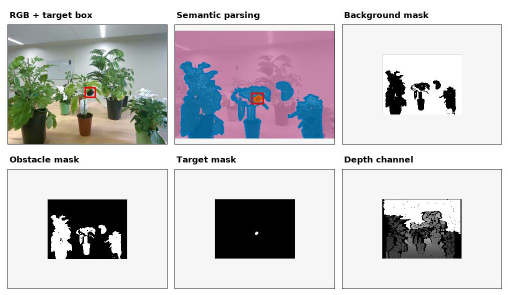}
    \caption{
    \textbf{Example TOB preprocessing outputs.}
    From left to right and top to bottom: raw RGB image with target box,
    semantic parsing result, background mask, obstacle mask, target mask, and
    normalized depth channel. The final policy input is formed by stacking the
    three semantic masks with the depth channel.
    }
    \label{fig:tob_preprocessing}
\end{figure}

Fig.~\ref{fig:tob_preprocessing} visualizes an example of the intermediate
outputs produced by the TOB preprocessing pipeline. Starting from the raw RGB
observation with a target box, the perception module generates a semantic
parsing result and then separates the scene into background, obstacle, and
target masks. These three masks are concatenated with the normalized depth
channel to form the four-channel TOB observation used by the policy. This
visualization illustrates how raw cluttered RGB-D observations are converted
into structured inputs that explicitly encode the target to reach, the obstacles
to avoid, and the background to ignore.

\end{document}